\documentclass[cameraready]{Interspeech}

\usepackage{amsmath,amssymb,amsfonts}
\usepackage{booktabs}
\usepackage{graphicx}
\usepackage{xcolor}
\usepackage{multirow}
\usepackage{hyperref}
\usepackage{algorithm}
\usepackage[noend]{algorithmic}

\newcommand{\lbs}{\ensuremath{B}}      %

\newcommand{\lepochs}{\ensuremath{E}}  %
\newcommand{\clientfrac}{\ensuremath{C}}    %

\newcommand{\loss}{\ell}
\newcommand{\SUB}[1]{\ENSURE \hspace{-0.15in} \textbf{#1}}

\newcommand{\nc}{K}
\newcommand{\pp}{\mathcal{P}}

\newcommand{\grad}{\triangledown}

\title{Federated Multilingual Speech-LLMs: Architecture and\\
       Aggregation Strategy Benchmarking}

\author[affiliation={1}, orcid=0000-0002-4507-4930,correspondingauthor]{Jordi}{Luque}
\author[affiliation={1,2}, orcid=0009-0007-7259-656X]{Aleix}{Sant}
\author[affiliation={1, 3}, orcid=0000-0002-7705-2250]{Fernando}{López}

\address{
    $^1$ Telefónica Innovación Digital, Scientific Research \\
    $^2$ Universitat Politècnica de Catalunya \\
    $^3$ Universidad Autónoma de Madrid
}

\email{jordi.luque@telefonica.com}

\keywords{federated learning, speech recognition, speech large language models, FedAvg, FedProx, multilingual ASR}
\begin{document}
\maketitle

\begin{abstract}
We present a comprehensive benchmark of Federated Learning (FL) for multilingual Automatic Speech Recognition (ASR), evaluating four Speech-LLM architectures on the Multilingual LibriSpeech dataset. We compare FedAvg and FedProx across frozen and unfrozen encoder configurations, demonstrating that optimized learning rates are critical for performance. Specifically, independently tuning the learning rates for the speech encoder, connector, and decoder yields the lowest error rates, with full three-component adaptation (LoRA for encoder and decoder, full training for the connector) producing the best FL results. We observe that FedProx efficacy is architecture-dependent, providing notable advantages in multilingual pre-trained architectures (e.g., EuroLLM over TinyLlama when keeping the encoder fixed); this indicates that LLM backbone capacity plays a key role in mediating resilience to heterogeneous data distributions. These findings offer concrete design guidance for deploying multilingual Speech-LLMs in privacy-sensitive, distributed environments.


\end{abstract}

\section{Introduction}
\label{sec:intro}
Speech large language models (LLMs) have emerged as a prominent paradigm for end-to-end automatic speech recognition (ASR), combining powerful acoustic encoders~\cite{radford2023whisper,chen2022wavlm} with pre-trained language decoders~\cite{zhang2023speechgpt,martins2024eurollm} through a lightweight cross-modal connector~\cite{tang2024salmonn}. This architecture leverages large-scale text pretraining to exploit rich language priors, enabling instruction-following speech understanding and achieving state-of-the-art results particularly in multilingual and low-resource settings. Privacy-sensitive deployments in healthcare, legal, and personal assistant applications motivate \emph{federated learning} (FL)~\cite{mcmahan2017communication}, where models are fine-tuned on distributed client data without centralising raw audio. 

Federated speech learning predates Speech-LLMs and has mainly targeted ASR, where strongly heterogeneous, privacy-sensitive speech data have motivated methods for non-IID optimisation, personalisation, and communication-efficient training~\cite{zhu22b_interspeech, applepfl4asr, 9747161}. More recent work consolidates parameter-efficient fine-tuning for pretrained speech encoders in FL, showing that LoRA and adapter-based approaches can effectively adapt foundation models under tight communication and compute budgets~\cite{du2024communication, kan2024parameter, ali2025eflpeft}. However, these studies operate on monolithic  speech encoder-only~\cite{kan2024parameter} or language-based architectures~\cite{applepfl4asr}, and no previous available work has directly investigated federated training of Speech-LLMs, leaving open how encoder, connector, and language-decoder components should be coordinated under multilingual, non-IID client distributions.


Motivated by these gaps, we focus explicitly on Speech-LLMs under the FL paradigm and benchmark four encoder--LLM families and two aggregation strategies in a realistic multilingual setting. Our study addresses the following questions: \vspace{-1mm}
\begin{itemize}
  \item \textbf{(Q1) Architecture}: which encoder--LLM combination best handles multilingual non-IID client distributions. Does the encoder type (ASR-supervised vs.\ self-supervised) affect federated adaptability of Speech-LLMs?
  \item \textbf{(Q2) Aggregation}: does FedProx mitigate client language drift in this multilingual Speech-LLM setting? 
\end{itemize}





These questions are particularly acute in multilingual settings, where clients differ simultaneously in language, accent, speaker population and recording conditions. Such strongly non-IID data cause \emph{client drift}~\cite{karimireddy2020scaffold} that destabilises FL optimisation~\cite{kairouz2021advances} and has motivated aggregation strategies such as FedProx~\cite{li2020fedprox} or SCAFFOLD~\cite{zhao2018federated}. 


Our FL experiments use a speaker-partitioned split of Multilingual LibriSpeech~\cite{pratap2020mls} (8 European languages, 316 clients, one speaker per client), inducing simultaneous acoustic, linguistic and domain-level heterogeneity. Local Speech-LLMs are fine-tuned with LoRA~\cite{hu2022lora} under frozen and unfrozen speech encoders, with the connector and LLM always unfrozen. We analyse the interplay between pretrained encoders and LLM backbones, the effect of differential per-component learning rates, and the FL-ASR behaviour of FedAvg and FedProx in this heterogeneous setting.




\section{System Architectures}
\label{sec:arch}

\emph{Speech-LLMs} couple three heterogeneous components: an acoustic encoder
($\mathcal{E}$), a cross-modal connector ($\mathcal{C}$), and a language
decoder ($\mathcal{L}$), producing a transcript as:

\begin{equation}
  \hat{\mathbf{y}} = \mathcal{L}\left([\mathcal{C}(\mathcal{E}(\mathbf{x}));\,\mathbf{E}_{\mathrm{text}}]\right).
  \label{eq:model}
\end{equation}

\begin{figure}[t]
  \centering
  \vspace{-0.4cm}
  \includegraphics[width=1\linewidth]{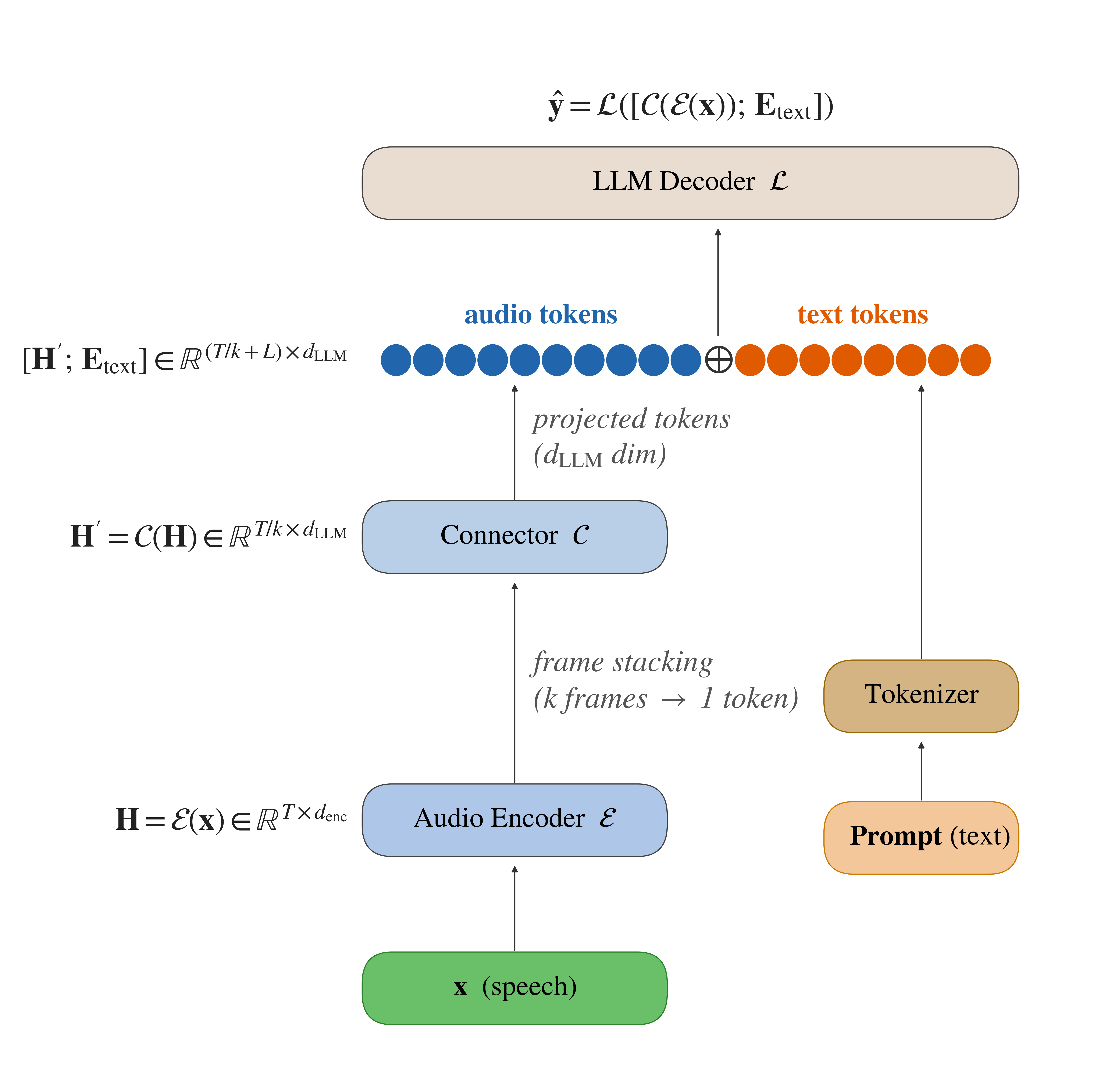}
  \caption{Speech-LLM architecture: encoder $\mathcal{E}$ maps speech $\mathbf{x}$ to frame-level features; connector $\mathcal{C}$ downsamples by $k$; decoder $\mathcal{L}$ generates $\hat{\mathbf{y}}$ from the projected tokens and task prompt.}
  \vspace{-0.3cm}
  \label{fig:arch}
\end{figure}

All three components are trained jointly by minimising the standard
autoregressive cross-entropy loss over the ground-truth transcript tokens
$\mathbf{y} = (y_1, \ldots, y_S)$:
\begin{equation}
  \mathcal{L}_{\mathrm{CE}}(\boldsymbol{\theta}) =
  -\sum_{s=1}^{S} \log p_{\boldsymbol{\theta}}\left(y_s \mid
    \mathcal{C}(\mathcal{E}(\mathbf{x})),\,
    \mathbf{E}_{\mathrm{text}},\,
    y_{<s}\right),
  \label{eq:loss}
\end{equation}
where $\boldsymbol{\theta}$ collects all trainable parameters (LoRA
adapters and connector projection), and $y_{<s}$ denotes the preceding
tokens supplied via teacher forcing.  The loss is computed over
transcript tokens; audio tokens and prompt embeddings appear as
conditioning context.  Figure~\ref{fig:arch} summarises the full pipeline.

\subsection{Encoder--LLM Models}
\label{sec:models}

The combination of foundational acoustic encoders and pretrained LLMs has become a prominent recipe for multilingual end-to-end speech recognition~\cite{zhang2023speechgpt}. 
We evaluate four encoder–LLM pairings:

\textbf{Whisper\,+\,TinyLlama.}
Whisper large-v3-turbo~\cite{radford2023whisper} encodes audio into
768-dimensional speech frame representations, projected to $2048$-dimensional tokens to align with TinyLlama-1.1B~\cite{zhang2024tinyllama} ($22$ layers). Whisper is a pretrained end-to-end multilingual ASR model, and its encoder ($32$ layers) already captures robust cross-lingual acoustic features.


\textbf{Whisper\,+\,EuroLLM.}
Identical Whisper encoder, but the decoder is replaced by EuroLLM-1.7B-Instruct~\cite{martins2024eurollm}, a decoder-only model pretrained on multilingual European text corpora.  

\textbf{WavLM\,+\,TinyLlama.}
WavLM-Large~\cite{chen2022wavlm} replaces Whisper as the encoder ($1024$-dimensional tokens, 24 layers); TinyLlama remains the decoder. WavLM is a self-supervised learning (SSL) model pretrained on masked speech prediction. Unlike Whisper it has \emph{never} seen ASR
supervision, making it a stronger test of whether FL can adapt an SSL encoder to the downstream ASR task.

\textbf{Voxtral-Mini}
Voxtral~\cite{mistral2025voxtral} is an end-to-end multimodal Speech-LLM composed of a Whisper-large-v3-based audio encoder and a 30-layer Ministral-3B~\cite{liu2026ministral3} text decoder, jointly pretrained on audio understanding and ASR. 


\subsubsection{Cross-modal connector.}

The connector $\mathcal{C}$, see Fig.~\ref{fig:arch}, bridges the acoustic encoder and the LLM
decoder in two steps.  First, a \emph{frame-stacking} operation with
stride $2$ concatenates each pair of consecutive encoder output
frames.  Second, a single
trainable linear layer projects to the input LLM dimension,
aligning acoustic representations with the LLM's embedding space. 
This design is shared by Whisper$+$TinyLlama, Whisper$+$EuroLLM, and
WavLM$+$TinyLlama, where only the single linear projection is trained from scratch. In contrast, Voxtral uses a different connector, downsampling the audio by a factor of~4, followed by a 2-layer MLP projector; a deeper, higher-compression connector in which both MLP layers are trained.


\section{Federated Setup}
\label{sec:fedsetup}

\begin{algorithm}[t]
\begin{algorithmic}
\SUB{Server:}
  \STATE initialise $\boldsymbol{\theta}_0$
  \FOR{each round $t = 1, 2, \dots, T$}
    \STATE $S_t \leftarrow$ random set of $\max(\clientfrac \cdot \nc,\,1)$ clients
    \FOR{each client $k \in S_t$ \textbf{in parallel}}
      \STATE $\boldsymbol{\theta}_{t}^{k} \leftarrow \text{ClientUpdate}(k,\boldsymbol{\theta}_{t-1})$
    \ENDFOR
    \STATE $\boldsymbol{\theta}_{t} \leftarrow \sum_{k \in S_t} \frac{n_k}{\sum_{j \in S_t} n_j}\boldsymbol{\theta}_{t}^{k}$
  \ENDFOR
  \STATE
\SUB{ClientUpdate($k$,$\boldsymbol{\theta}$):}
  \FOR{local epoch $e = 1, \dots, \lepochs$}
    \FOR{each batch $b \subseteq \pp_k$, $|b|{=}\lbs$}
      \STATE $\boldsymbol{\theta} \leftarrow \mathrm{AdamW}(\boldsymbol{\theta},\;\grad\loss(\boldsymbol{\theta};\,b),\;\eta)$
    \ENDFOR
  \ENDFOR
  \STATE \textbf{return} $\boldsymbol{\theta}$
\end{algorithmic}

\caption{FedAvg~\cite{mcmahan2017communication}. $\boldsymbol{\theta}$: LoRA adapters + connector; $K$: clients; $n_k$: samples on client $k$; $\clientfrac$: client fraction/round; $\lbs$: batch size; $\lepochs{=}10$: local epochs; $\eta{=}10^{-4}$.}
\label{alg:fedavg}
\end{algorithm}

\subsection{Aggregation Strategies}
\label{sec:aggregation}
Our federated setting uses synchronous client-server optimization where participating clients ($C=0.3$) perform $E=10$ local AdamW epochs before server aggregation via FedAvg or FedProx. We compute the loss gradient over all data held by these clients, $C*K=94$ clients for the global server; see Algorithm~\ref{alg:fedavg}. Two aggregation strategies are considered:

\textbf{FedAvg}: Standard weighted averaging of client updates~\cite{mcmahan2017communication} or each round:
\begin{equation}
  \boldsymbol{\theta}^{(t+1)} =
  \sum_{k \in S_t} \frac{n_k}{\sum_{j \in S_t} n_j}
  \boldsymbol{\theta}^{(t,k)},
  \label{eq:fedavg}
\end{equation}
where $n_k$ are the training samples on client $k$ and $S_t$ is the random fraction of clients sampled every round. 

\textbf{FedProx}: Regularises the client loss, \emph{ClientUpdate} in Algorithm~\ref{alg:fedavg}, with the proximal term $\mu$ to limit client drift~\cite{li2020fedprox}, due to data heterogeneity, for each round:
\begin{equation}
  \min_{\boldsymbol{\theta}} 
  \mathcal{L}_{\text{local}}(\boldsymbol{\theta})
  + \frac{\mu}{2}\left\lVert\boldsymbol{\theta}
    - \boldsymbol{\theta}^{(t)}\right\lVert_2^2,
  \label{eq:fedprox}
\end{equation}
where $\mu{>}0$ penalises deviation, e.g. due to narrow acoustic distribution in client data, from the current global model $\boldsymbol{\theta}^{(t)}$. 



\subsection{Dataset and Partitioning}
\label{sec:data}

\subsubsection{Corpus}
All experiments use the Multilingual LibriSpeech (MLS)
corpus~\cite{pratap2020mls}, an audiobook corpus covering 8 European languages derived from LibriVox recordings.  We use the official MLS \texttt{train} splits as the federated training pool (685.7~h total),
the MLS \texttt{dev} split for validation during training, and the MLS \texttt{test} split as the held-out evaluation benchmark (138~h, 19,492 samples). Table~\ref{tab:data} reports training hours and speaker counts per language.

\subsubsection{Multilingual Partition (approximately IID)}
\label{sec:a1partition}

The multilingual partition serves as a control to isolate the effect of data heterogeneity from architecture choice. A random multilingual mixture of utterances is assigned to each client irrespective of speaker identity, approximating the IID assumption. Because MLS is derived from LibriVox audiobooks, the same reader can appear across official train, dev, and test splits. The centralized training pools this full set and therefore carries a 3.5\% speaker leakage. For FL experiments , partition~(A) in Table~\ref{tab:wer}, same identities overlap contaminating 59 clients out of $K{=}316$ clients (18.7\%). These clients contain utterances from test or dev sharing same speakers, totalling 60,825 of 1,726,583 training samples (i.e the same 3.5\% speaker leakage). Note that these are different utterances of the same speaker, not duplicate audio.

\subsubsection{Speaker Partition (non-IID)}
\label{sec:b1partition}

The speaker partition, (B) in Table~\ref{tab:wer}, assigns all utterances of a single MLS speaker to one client. With $K{=}316$ clients, this creates the strongest possible non-IID distribution: each client's data is drawn from a single acoustic identity, language, and recording environment, producing simultaneous \emph{linguistic} (each client speaks at most one language) and \emph{acoustic} (microphone, room, speaking rate) heterogeneity. 
Due to the LibriVox origin of MLS, 8 of 316 clients (2.5\%) correspond to speakers also present in the MLS test split, accounting for 4,747 of 169,586 training samples (2.8\%), while 3.5\% of centralized training data and 18.7\% of clients in partition (A) share speakers with the test set. Because these overlap rates are inherited from the standard MLS splits, the centralized and IID upper bounds benefit from acoustic speaker familiarity. Consequently, the actual degradation caused by FL non-IID conditions is slightly less severe than the raw distance to the upper bounds suggests.

\begin{table}[t]
  \centering
  \caption{MLS training data by language under the stratified by speaker partition, 316 clients.}
  \label{tab:data}
  \setlength{\tabcolsep}{4pt}
  \begin{tabular}{lrr}
    \toprule
    \textbf{Language} & \textbf{Hours} & \textbf{Speakers/Clients} \\
    \midrule
    French     & 251.6 & 15 \\
    German     & 160.9 & 19 \\
    English    & 105.3 & 256 \\
    Spanish    &  83.8 &  9 \\
    Italian    &  27.3 &  7 \\
    Polish     &  25.7 &  1 \\
    Portuguese &  18.5 &  5 \\
    Dutch      &  12.7 &  4 \\
    \midrule
    \textbf{Total} & \textbf{685.7} & \textbf{316} \\
    \bottomrule
  \end{tabular}
\end{table}

\subsection{Training Configuration}
\label{sec:trainconfig}

All FL experiments use the Flower simulation
framework with Ray as the backend ~\cite{beutel2020flower}.  
Experiments on the non-IID speaker partition (B) train for $T=40$ global rounds. For the multilingual partition (A), results are reported at $T=9$ global rounds ($\ddagger$ in Table 2). This is because the multilingual mixture in partition (A) exhibits rapid convergence leading to early WER optimization collapse. All hyper-parameter choices---encoder learning-rate multiplier $\lambda$, FedProx coefficient $\mu$, and local epochs $E$---were selected on the MLS dev set, and the test split is used only to evaluate each selected checkpoint. Local clients perform $E{=}10$ local epochs per round, using AdamW optimiser with maximum learning rate $\eta{=}10^{-4}$, cosine decay, batch size 16, and half precision bf16. Centralized (non-FL) training uses the same AdamW optimiser with cosine decay over a maximum of $10$ epochs; the best-validation-WER checkpoint is selected (typically around epoch~$4$). Evaluation uses the MLS test split with overall WER reported across all 8 languages combined. All models are fine-tuned with LoRA~\cite{hu2022lora} adapters (rank $r{=}8$, $\alpha{=}16$, dropout $0.05$; Voxtral uses $\alpha{=}32$) applied to the attention projections of the speech encoder ($q,k,v$) and LLM decoder ($q,v$), together with the fully-trained connector; only these parameters are trainable and are transmitted between clients and the FL server, reducing communication cost over full fine-tuning. Unlike the pretrained encoders and LLM decoders, we initialise the connector from scratch (except for Voxtral), so it must learn to bridge the modalities entirely from the federated fine-tuning data. Unless stated otherwise, WER values are reported as proportions (e.g., $0.1415$ corresponds to $14.15\%$).

\vspace{-.10cm}
\section{Results and Discussion}
\label{sec:results}

\begin{table}[t]
  \centering
  \caption{WER on MLS test set. B: non-IID speaker partition. A: approximately IID multilingual partition. Gap: WER relative to corresponding FedAvg baseline; negative = improvement. $\dagger$: best encoder learning rate multiplier ($0.02\times$LLM lr). $\ddagger$: reported at round~9.}
  \label{tab:wer}
  \setlength{\tabcolsep}{4pt}
  \begin{tabular}{llrr}
    \toprule
    \textbf{System} & \textbf{Encoder Setting} & \textbf{WER} & \textbf{Gap} \\
    \midrule
    \multicolumn{4}{l}{\textit{Whisper\,+\,TinyLlama-1.1B}} \\
    Centralized & frozen          & 0.0719 & ---    \\
    FedAvg      & frozen (B)     & 0.1415 & ---    \\
    FedAvg      & frozen (A)$^\ddagger$ & 0.1284 & $-0.0131$ \\
    FedAvg      & unfrozen (B)$^\dagger$ & 0.1418 & $+0.0003$ \\
    FedProx $\mu{=}0.001$ & frozen (B) & 0.1499 & $+0.0084$ \\
    FedProx $\mu{=}0.001$ & frozen (A)$^\ddagger$ & 0.1386 & $+0.0102$ \\
    FedProx $\mu{=}0.05$  & frozen (B) & 0.1615 & $+0.0200$ \\
    FedProx $\mu{=}0.1$   & frozen (B) & 0.1843 & $+0.0428$ \\
    \midrule
    \multicolumn{4}{l}{\textit{Whisper\,+\,EuroLLM-1.7B-Instruct}} \\
    Centralized & frozen         & 0.0660 & ---    \\
    FedAvg      & frozen (B)     & 0.1330 & ---    \\
    FedAvg      & frozen (A)$^\ddagger$ & 0.1224 & $-0.0106$ \\
    FedAvg      & unfrozen (B)   & 0.1778 & $+0.0448$ \\
    FedAvg      & unfrozen$^\dagger$ (B) & \textbf{0.1170} & {$\mathbf{-0.0160}$} \\
    FedProx $\mu{=}0.001$ & frozen (B) & 0.1217 & $-0.0113$ \\
    \midrule
    \multicolumn{4}{l}{\textit{WavLM-Large\,+\,TinyLlama-1.1B}} \\
    Centralized & frozen         & 0.2409 & ---    \\
    FedAvg      & frozen (B)    & 0.5559 & ---    \\
    FedAvg      & unfrozen (B)  & 0.5338 & $-0.0221$ \\
    \midrule
    \multicolumn{4}{l}{\textit{Voxtral-Mini-3B-2507}} \\
    Centralized & frozen & 0.1238 & ---    \\
    FedAvg      & frozen (B)    & 0.1442 & ---    \\
    FedAvg      & unfrozen (B)  & 0.1362 & $-0.0080$ \\
    FedProx $\mu{=}0.001$ & frozen (B)   & 0.1492 & $+0.0050$ \\
    FedProx $\mu{=}0.001$ & unfrozen (B) & 0.1468 & $+0.0106$ \\
    \bottomrule
  \end{tabular}
\end{table}

\subsection{Centralized and IID Upper Bounds}
\label{sec:centralized}

Centralized training (non-FL) on the pooled partition sets the per-architecture performance ceiling (Table~\ref{tab:wer}); all centralized models converge within $4$--$7$ epochs. Whisper$+$EuroLLM reaches WER $0.066$ versus $0.072$ for Whisper$+$TinyLlama, confirming the multilingual LLM advantage even without non-IID pressure; WavLM$+$TinyLlama reaches $0.24$ ($3.3\times$ higher than the Whisper variant), showing that ASR pretraining of the encoder decisively beats SSL pretraining for ASR. 
\vspace{-.22cm}
\subsection{Aggregation Strategy Comparison}
\label{sec:fedavg}
With a frozen encoder, FedAvg reaches WER $0.1415$ ($96.8\%$ relative degradation
from centralized) for Whisper$+$TinyLlama; replacing only the decoder with EuroLLM reduces it to $0.133$ ($-6.0\%$ relative), and Voxtral ($0.144$) underperforms the EuroLLM pipeline. WavLM reaches $0.56$, confirming that ASR-pretrained encoders are decisive under FL non-IID conditions. FedProx regularisation degrades Whisper$+$TinyLlama monotonically with $\mu$, and also hurts Voxtral (frozen $0.144\to0.149$, unfrozen $0.1362\to0.1468$), contradicting its non-IID motivation.
Strikingly, FedProx $\mu{=}0.001$ improves Whisper$+$EuroLLM (WER $\mathbf{0.1217}$, $-8.5\%$ relative), the best frozen-encoder result. 
\begin{figure}[t]
  \centering
  \vspace{-0.4cm}
  \includegraphics[width=\linewidth]{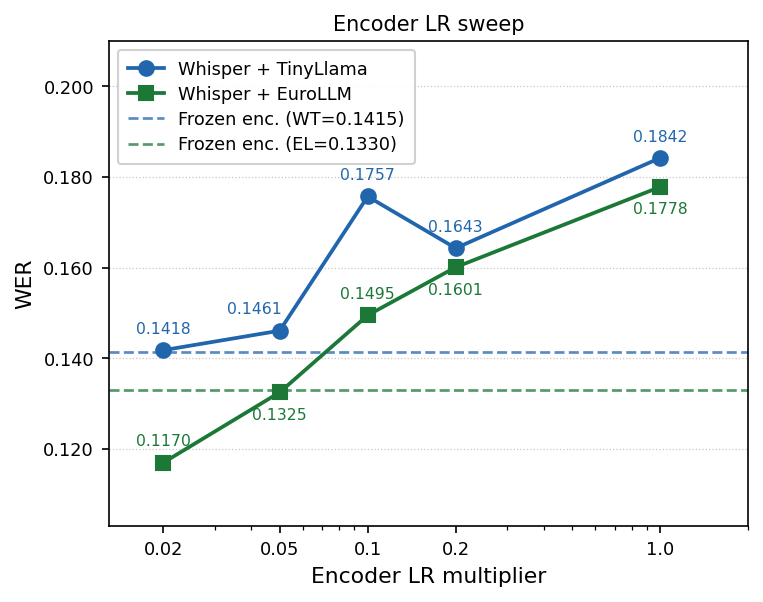}
  \caption{WER vs.\ encoder learning-rate multiplier (relative to LLM LR). Whisper$+$TinyLlama (blue, $E{=}5$) stays flat at its frozen baseline while Whisper$+$EuroLLM (green, $E{=}10$) improves monotonically; dashed: frozen-encoder baselines.}
  \vspace{-0.3cm}
  \label{fig:enc-lr-sweep}
\end{figure}
\vspace{-.20cm}
\subsection{Encoder LR Tuning under FL}
\label{sec:encoder-unfrozen}
Unfreezing the Whisper encoder under FedAvg yields WER $0.1418$, comparable to the frozen baseline ($0.1415$), even after sweeping the encoder learning rate $\in \{0.01,0.02,0.05,0.10\}$ and for different values of local epochs $E\in\{3,5,10\}$, see Fig.~\ref{fig:enc-lr-sweep}. We hypothesise that this stems from encoder-update cancellation, in which each client adapts the encoder to its own speaker's acoustics, producing divergent updates that average to near-zero under FedAvg, while LLM updates on the shared linguistic task survive averaging.  WavLM confirms the pattern. The centralized gap ($3.2\times$) grows to $3.9\times$ under FL ($0.24\to0.56$), suggesting SSL pretraining misalignment is amplified by heterogeneous data. Unfreezing the WavLM encoder provides only marginal gain. 
For EuroLLM, after sweeping the learning rate multiplier for the encoder we can observe a monotonic improvement from $0.178\to\mathbf{0.117}$, see Fig.~\ref{fig:enc-lr-sweep}, a $12\%$ relative below the encoder-frozen EuroLLM variant. Voxtral also benefits modestly from unfreezing ($0.144\to0.136$). 
These results suggest that, for multicomponent Speech-LLMs, separately adjusting the encoder and decoder is crucial when the encoder is well pre-trained, in order to prevent update cancellation during federated aggregation.

\vspace{-.20cm}
\subsection{Per-Language Analysis}
\label{sec:perlang}

Table~\ref{tab:perlang} compares FedAvg TinyLlama and EuroLLM against the FedProx with $\mu{=}0.001$ which reaches the lower WER $0.122$ between frozen-encoder FL results. For the FedAvg algorithm, EuroLLM also outperforms TinyLlama on six of eight languages; with the largest gains are on most of low-resource languages, see Table~\ref{tab:data}, where multilingual priors matter most: Portuguese ($-6.5\%$ abs.), Italian ($-4.8\%$), Polish ($-3.3\%$).  
FedProx EuroLLM improves over FedAvg on five of eight languages: Dutch ($0.265\to0.195$, $-7.0\%$ abs.), Polish, English, French, and Spanish. Conversely, it degrades German ($+2.6\%$ abs.), Italian, and Portuguese. Note that benefit is strongest for the least-represented languages ($\leq4$ clients), while moderate-resource languages show mixed results.

Because the client distribution is strongly skewed toward English (256 of 316 clients) and the overall WER is dominated by high-resource evaluation words, we also report macro-averaged WER (unweighted mean of the per-language values in Table~\ref{tab:perlang}): FedAvg Whisper$+$TinyLlama is $0.170$ versus $0.150$ for Whisper$+$EuroLLM. This macro gap ($-2.0$ points) is larger than the word-weighted gap ($-0.9$ points), confirming that the EuroLLM advantage is driven particularly by the low-resource languages rather than being masked by them.

\begin{table}[t]
  \centering
  \caption{Per-language WER on MLS test set for FL encoder frozen variants trained on speaker (non-IID) partition. Cent.\ = Centralized Whisper$+$TinyLlama. FedAvg\textsubscript{WT} = FedAvg Whisper$+$TinyLlama. FedAvg\textsubscript{EL} = FedAvg Whisper$+$EuroLLM. FP\textsubscript{EL} = FedProx $\mu{=}0.001$ Whisper$+$EuroLLM. Bold: best FL result per language.}
  \label{tab:perlang}
  \setlength{\tabcolsep}{3pt}
  \small
  \begin{tabular}{lcccc}
    \toprule
    \textbf{Lang.} &
    \textbf{Cent.} &
    \textbf{FedAvg\textsubscript{WT}} &
    \textbf{FedAvg\textsubscript{EL}} &
    \textbf{FP\textsubscript{EL}} \\
    \midrule
    Dutch      & 0.110 & 0.270 & 0.265 & \textbf{0.195} \\
    English    & 0.049 & 0.100 & 0.098 & \textbf{0.090} \\
    French     & 0.061 & 0.075 & 0.078 & \textbf{0.072} \\
    German     & 0.071 & 0.104 & \textbf{0.098} & 0.125 \\
    Italian    & 0.121 & 0.233 & \textbf{0.186} & 0.189 \\
    Polish     & 0.110 & 0.270 & 0.237 & \textbf{0.228} \\
    Portuguese & 0.093 & 0.225 & \textbf{0.160} & 0.177 \\
    Spanish    & 0.046 & 0.085 & 0.075 & \textbf{0.068} \\
    \midrule
    \textbf{Overall} & \textbf{0.072} & 0.142 & 0.133 & \textbf{0.122} \\
    \bottomrule
  \end{tabular}
\end{table}

\vspace{-.20cm}
\section{Conclusion}
\label{sec:conclusion}

This work benchmarks federated training for Speech-LLMs across four architectures and two aggregation strategies using a partition of 316 clients from the Multilingual LibriSpeech dataset. The results show that multilingual pre-training is a primary driver of FL performance, with EuroLLM outperforming TinyLlama, by 6\% relative WER, and Voxtral pipelines. Furthermore, the efficacy of the FedProx algorithm is architecture-dependent, improving Whisper+EuroLLM while degrading Whisper+TinyLlama and Voxtral configurations. Controlled paired comparisons suggest that stronger multilingual LLM capacity helps absorb client drift under proximal regularization, though broader multi-component variations also introduce secondary effects from connector depth and encoder pre-training. We also report that differential learning rate tuning (low-rate encoder vs.\ high-rate decoder) provides superior results compared to frozen encoders, since the components of a Speech-LLM exhibit distinct gradient scales and adaptation requirements that preclude treating the model as monolithic during FL optimisation. These results offer a road map for practitioners deploying robust, cross-lingual Speech-LLM systems in federated settings.
%
\section{Acknowledgements}
This work has received funding from the European Union's Horizon Europe research and innovation programme under the project ELOQUENCE (Grant Agreement No. 101135916). This work was supported by computational resources from the EuroHPC Joint Undertaking under the EuroHPC AI Factory grant EHPC-AIF-2026LS01-004.

\bibliographystyle{IEEEtran}
\bibliography{fl_benchmark}

\end{document}